\documentclass[10pt,twocolumn,letterpaper]{article}

\usepackage[pagenumbers]{cvpr} 

\definecolor{cvprblue}{rgb}{0.21,0.49,0.74}
\usepackage[pagebackref,breaklinks,colorlinks,allcolors=cvprblue]{hyperref}

\def\paperID{*****} 
\def\confName{CVPR}
\def\confYear{2026}

\title{Geometric Disentanglement of Text Embeddings \\ for Subject-Consistent Text-to-Image Generation using A Single Prompt}

\author{Shangxun Li\\
Yonsei University\\
Seoul, South Korea\\
{\tt\small lee.sanghoon@yonsei.ac.kr}
\and
Youngjung Uh\\
Yonsei University\\
Seoul, South Korea\\
{\tt\small yj.uh@yonsei.ac.kr}
}

\begin{document}
\twocolumn[{%
\renewcommand\twocolumn[1][]{#1}%
\maketitle
\begin{center}
    \includegraphics[width=\textwidth]{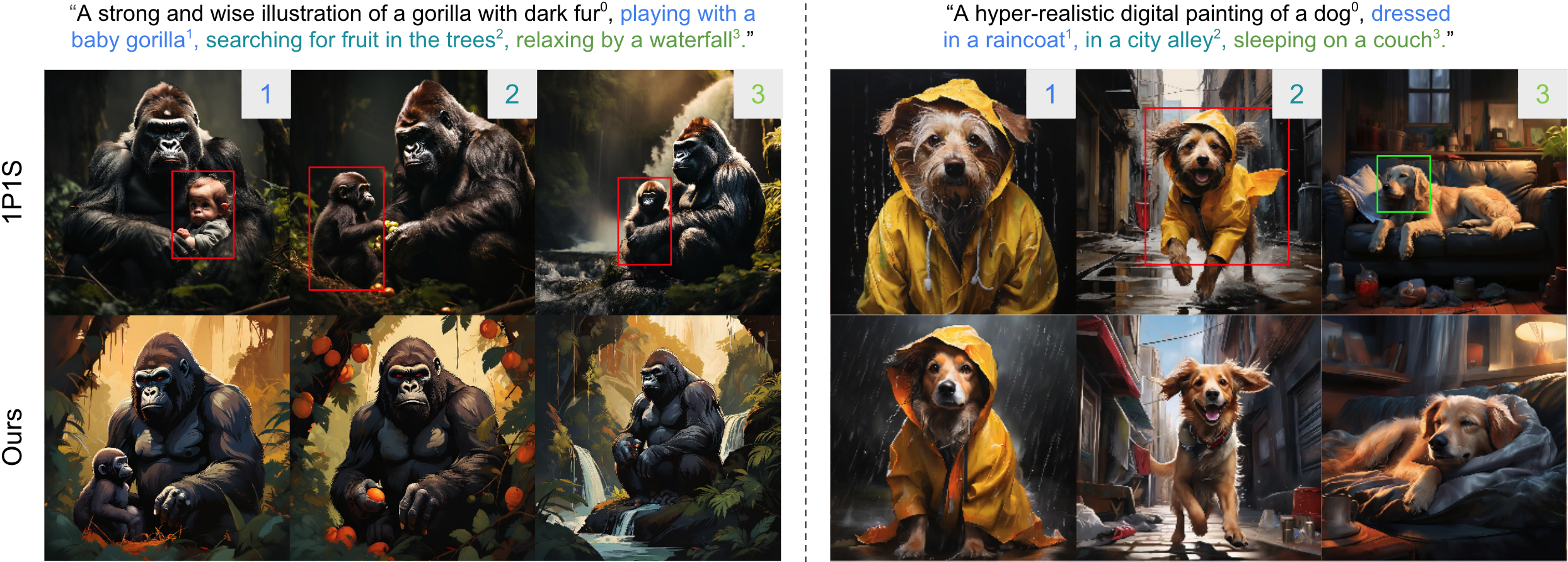} 
    \captionof{figure}{\textbf{Overview.} \textit{1Prompt1Story} exhibits severe text misalignment due to semantic leakage (marked in red), where concepts from preceding frames bleed into subsequent ones. For instance, the “baby gorilla” from the first frame reappears incorrectly in the following two frames, and the “raincoat” from the first dog image persists in the second. Furthermore, it suffers from noticeable subject inconsistency (marked in green), where the dog's breed changes entirely in the third frame. In contrast, our method demonstrates superior performance, maintaining strong subject consistency and precise text alignment across all generated images.}
    \label{fig:teaser}
\end{center}
}]

\begin{abstract}
Text-to-image diffusion models excel at generating high-quality images from natural language descriptions but often fail to preserve subject consistency across multiple outputs, limiting their use in visual storytelling. Existing approaches rely on model fine-tuning or image conditioning, which are computationally expensive and require per-subject optimization. \textit{1Prompt1Story}, a training-free approach, concatenates all scene descriptions into a single prompt and rescales token embeddings, but it suffers from semantic leakage, where embeddings across frames become entangled, causing text misalignment. In this paper, we propose a simple yet effective training-free approach that addresses semantic entanglement from a geometric perspective by refining text embeddings to suppress unwanted semantics. Extensive experiments prove that our approach significantly improves both subject consistency and text alignment over existing baselines.
\end{abstract}    
\section{Introduction}
\label{sec:intro}

In recent years, text-to-image diffusion models have achieved significant progress, enabling users to generate diverse, high-quality images from natural language descriptions \cite{podell2023sdxlimprovinglatentdiffusion, rombach2021highresolution}. Despite their advances, such models often struggle to maintain consistent depictions of subjects across a sequence of images, a capability crucial for visual storytelling.

A dominant approach to this challenge is personalization \cite{gal2022imageworthwordpersonalizing, ruiz2023dreamboothfinetuningtexttoimage}, where a text-to-image model is fine-tuned on user-provided images of a target subject. While effective, personalization requires time-consuming per-subject optimization and is prone to overfitting, where maintaining subject identity comes at the cost of text alignment. Alternatively, other methods incorporate image conditioning by training models with external image encoders \cite{gal2023encoderbaseddomaintuningfast, wei2023eliteencodingvisualconcepts, ye2023ipadaptertextcompatibleimage}, but these approaches demand significant computational costs.

To avoid such computational overhead, several recent works have investigated training-free strategies. \textit{StoryDiffusion} \cite{zhou2024storydiffusion} and \textit{ConsiStory} \cite{tewel2024training} modify the self-attention mechanism of a pre-trained diffusion model to enable feature sharing across images generated in a batch to ensure subjects remain consistent. However, these methods require substantial memory to store and reuse shared features during generation. Apart from these approaches, instead of feeding the model separate text prompts for each scene, \textit{1Prompt1Story} \cite{liu2025onepromptonestoryfreelunchconsistenttexttoimage} concatenates all scene descriptions into a single prompt that begins with an identity description specifying the subject’s visual attributes, followed by frame descriptions for each scene, leveraging language models’ contextual understanding capability to maintain subject consistency through shared context. Although this approach has shown plausible results, it suffers from semantic leakage, where token embeddings across frames become entangled, leading to text misalignment (see \cref{fig:teaser} top).

To overcome this limitation, we propose a simple yet effective training-free method that addresses semantic entanglement from a geometric perspective. Our approach builds on the single-prompt paradigm introduced by \textit{1Prompt1Story}. To purify the entangled semantics in the text embedding for individual scene generation, our approach refines the text embedding through a projection-based operation that maps it onto a subspace orthogonal to undesired token components, thereby attenuating unwanted semantics while preserving subject identity. Extensive evaluations demonstrate that our method significantly outperforms existing baselines in both subject consistency and text alignment.

\section{Method}
\label{sec:method}


\subsection{Consistent Text-to-Image Generation from A Single Prompt}

\begin{figure}[t]
  \centering
  \vspace{5mm}
  \includegraphics[width=1\linewidth]{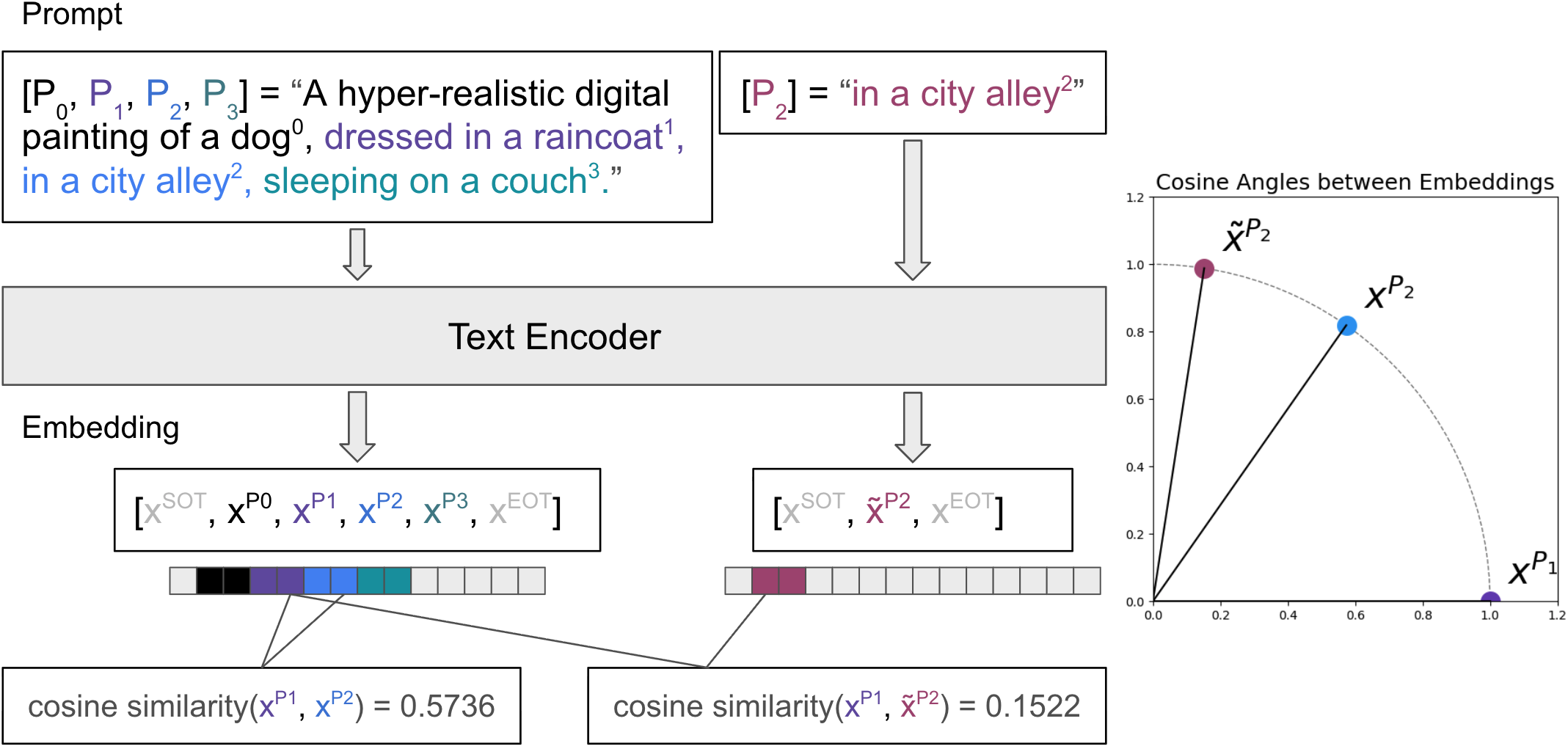}
  \caption{\textbf{An Illustration of Semantic Entanglement in Text Embeddings.} As shown on the left, the causal self-attention mechanism in the text encoder causes semantic information to flow from earlier parts of the prompt (e.g., $P_1$: “dressed in a raincoat”) to later parts (e.g., $P_2$: “in a city alley”). The cosine angles on the right quantify this entanglement, revealing a high cosine similarity (0.5736) between the embeddings $x^{P_1}$ and $x^{P_2}$.}
  \label{fig:cossim}
\end{figure}

Instead of providing the model with different text inputs for each frame, the single-prompt paradigm constructs a single text input $P = [P_0; P_1; ... ; P_N]$ for the model that starts with an initial identity prompt $P_0$, which describes the target subject’s visual attributes, followed by a series of frame prompts $P_i$ that specify different scenarios for each frame, implicitly maintaining identity consistency through shared context. This effect stems from the self-attention mechanism within Transformer-based text encoders \cite{vaswani2023attentionneed}, which allows semantic information to flow from earlier to later parts of the sequence. While this property helps maintain identity consistency as all the subsequent frame prompts $P_i$ attend to the prefixed identity prompt $P_0$, it also causes the semantics of individual frame prompts to become intertwined within the concatenated prompt embedding (see \cref{fig:cossim}).

\subsection{Orthogonal Semantic Subspace Projection}

To mitigate semantic entanglement, we leverage the structured semantic geometry of the CLIP \cite{radford2021learningtransferablevisualmodels} embedding space, whose locally linear properties enable the systematic manipulation of concepts through vector operations \cite{grand2018semanticprojectionrecoveringhuman, mikolov2013distributedrepresentationswordsphrases, mikolov-etal-2013-linguistic}. Specifically, we project the text embedding matrix onto the semantic subspace of an undesired concept and subtract this projection to suppress its influence.

\begin{figure*}[t!]
  \centering
  \vspace{5mm}
  \includegraphics[width=1\textwidth]{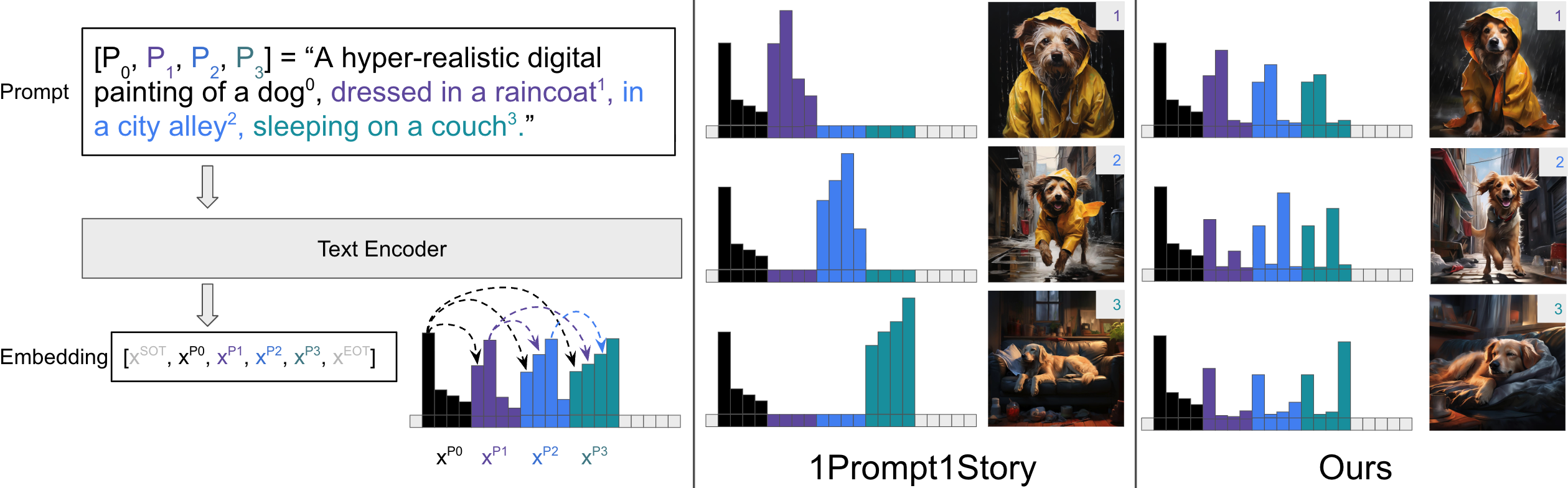}
  \caption{\textbf{An Illustration of Our Approach Compared to \textit{1Prompt1Story}.} The diagram on the left shows how a concatenated prompt is processed by a text encoder, where the causal self-attention mechanism induces semantic entanglement across frame prompt embeddings. The colored bars visualize these embeddings, with each dimension representing the semantic strength of concepts from $[P_0; P_1; P_2; P_3]$. \textit{1Prompt1Story} reweights embeddings by scaling up the current frame and downscaling others, but it suffers from semantic leakage, where entangled embeddings cause severe text misalignment, and subject inconsistency, where aggressive downscaling causes loss of identity information. In contrast, our method separates undesired semantics from the text embedding while preserving essential information for rendering the target subject.}
  \label{fig:comparison}
\end{figure*}

Given a concatenated prompt $P = [P_0; P_1; ... ; P_N]$, we partition it into two semantic sets: the express set $P^{exp}=[P_0; P_j]$, consisting of the identity and current frame prompts that must be preserved and accurately rendered in the generated image, and the suppress set $P^{sup} = \{P_k|k \in [1, N], k \neq j\}$, containing the remaining prompts to be suppressed. Using these sets, we decompose the full prompt embedding $X= \tau(P)$ into the express vector $E=XP_{X^{exp}}$ and the suppress vector $S=XP_{X^{sup}}$, obtained by projecting $X$ onto the subspaces spanned by $X^{exp}= \tau(P^{exp})$ and $X^{sup} = \tau(P^{sup})$ using the projection matrices $P_{X^{exp}}$ and $P_{X^{sup}}$, respectively, capturing desired and undesired semantics.

To separate the text embedding $X$ from the subspace spanned by $X^{sup}$, we remove the suppress vector $S$ from $X$ as $X^{'} = X - \alpha S$, where $\alpha \in [0, 1]$ controls the degree of suppression. However, directly subtracting $S$ can be overly aggressive, as it alters the embedding direction and risks distorting desired semantics, particularly when desired and undesired concepts are closely related, leading to inadvertent removal of semantic components essential for accurately rendering the desired concepts. To address this, we propose a dual-subspace approach that performs semantically-aware subtraction by first disentangling desired and undesired components. Specifically, we purify the suppress vector $S$ by projecting it orthogonally to the express vector $E$. This ensures that only semantics independent of the desired content are removed, thereby preserving the integrity of the target subject and scene. This is achieved by subtracting the projection of $S$ onto $E$ from $S$ itself so that the resulting purified suppress vector $S^{'}$ is orthogonal to $E$:
\begin{equation}
  \ S' = S - \frac{S \cdot E}{||E||^{2}} \cdot E
  \label{eq:main_func}
\end{equation}
Finally, the purified suppress vector $S^{'}$ is is scaled by the hyperparameter $\alpha$ and subtracted from the original full embedding $X$ to produce the refined embedding $X^{'} = X - \alpha S^{'}$. By subtracting the purified component $S^{'}$ instead of the original suppress vector $S$, our approach ensures that the suppression process does not inadvertently degrade the core concepts defined in the express set.

To obtain the projection matrices $P_{X^{exp}}$ and $P_{X^{sup}}$, we employ SVD, where it is known that the projection matrix $P_{\tilde{X}}$ can be defined using the orthonormal matrix $\tilde{V}$ as $P_{\tilde{X}} = \tilde{V}\tilde{V}^{T}$, where $\tilde{X}=\tilde{U}\tilde{\Sigma}\tilde{V}^{T}$ is the singular value decomposition of $\tilde{X}$, $\tilde{X}$ is the embedding matrix of the concepts we aim to express or suppress, and $P_{\tilde{X}}$ is the projection matrix onto the subspace spanned by $\tilde{X}$.

\section{Experiment}

To validate the effectiveness of our method, we conduct experiments comparing its performance against established baselines. We evaluate two key aspects of consistent text-to-image generation: the model’s ability to maintain subject identity across multiple frames and its fidelity to the specific textual description for each frame.

\subsection{Experiment Settings}

Following the evaluation protocol of \textit{1Prompt1Story}, we conduct our experiments on the  \textit{ConsiStory+} benchmark, which consists of 192 distinct prompt sets, from which we generated up to 1100 images for evaluation.

\subsection{Evaluation Metrics}
To evaluate text alignment, we compute the average CLIP-Score \cite{hessel2022clipscorereferencefreeevaluationmetric} between each generated image and its corresponding prompt, denoted as CLIP-T. For subject consistency, we measure image similarity using DreamSim \cite{fu2023dreamsim}, a validated proxy for human judgment of visual similarity, along with CLIP-I, defined as the cosine similarity between image embeddings, and DINO \cite{caron2021emergingpropertiesselfsupervisedvision, oquab2023dinov2}, a feature similarity metric based on DINO embeddings of the generated images.

\begin{table}[t] 
\centering
\resizebox{\columnwidth}{!}{%
\begin{tabular}{lcccc}
\toprule
Method & CLIP-T $\uparrow$ & CLIP-I $\uparrow$ & DINO $\uparrow$ & DreamSim $\downarrow$ \\
\midrule
SDXL & \textbf{0.8889} & 0.8972 & 0.7298 & 0.1989 \\
1P1S & 0.8252 & 0.8826 & 0.6898 & 0.2195 \\
Ours & 0.8766 & \textbf{0.9168} & \textbf{0.7900} & \textbf{0.1570} \\
\bottomrule
\end{tabular}%
} 
\caption{\textbf{Quantitative Comparisons.}}
\label{tab:quantitative_comparisons}
\end{table}

\subsection{Quantitative Analysis}
The quantitative results of our comparative evaluation are summarized in \cref{tab:quantitative_comparisons}. Our method achieves state-of-the-art performance across all three subject consistency metrics while maintaining text alignment close to vanilla SDXL. As expected, SDXL reports the highest CLIP-T score, serving as an upper bound since it is not constrained by identity preservation tasks. 1Prompt1Story exhibits a significant decline in text alignment, confirming our claim that its mechanism is prone to semantic leakage and text misalignment. Our approach attains a CLIP-T score of 0.8766. This result is remarkably close to the unconstrained baseline, demonstrating that our approach is highly effective at preserving the desired semantics of the target frame while suppressing the undesired ones, thus preventing text misalignment.

\subsection{Ablation Study}

The comparison between single- and dual-subspace methods validates our hypothesis that explicitly disentangling and preserving express semantics during suppression allows the dual-subspace method to effectively remove undesired semantics without discarding components essential for rendering the target concepts (\cref{tab:ablation_study}). We also test a variant that scales down undesired token embeddings after our dual-subspace approach. While this achieves a CLIP-T score comparable to our full method, it consistently degrades subject consistency, as indiscriminate suppression removes semantics crucial to the target subject. These results confirm that explicitly protecting express semantics is key to maintaining subject consistency.

\begin{table}[h]
\centering
\resizebox{\columnwidth}{!}{%
\begin{tabular}{lcccc}
\toprule
Method & CLIP-T $\uparrow$ & CLIP-I $\uparrow$ & DINO $\uparrow$ & DreamSim $\downarrow$ \\
\midrule
Single & 0.8592 & 0.8879 & 0.7505 & 0.1836 \\
Dual w/ rescale & 0.8765 & 0.9148 & 0.7847 & 0.1594 \\
Dual (Ours) & \textbf{0.8766} & \textbf{0.9168} & \textbf{0.7900} & \textbf{0.1570} \\
\bottomrule
\end{tabular}%
} 
\caption{\textbf{Ablation Study.} ``Single'' and ``Dual'' represent single- and dual-subspace respectively.}
\label{tab:ablation_study}
\end{table}
\section{Conclusion}

In this paper, we addressed the challenge of semantic entanglement in consistent text-to-image generation from a single prompt. We proposed a novel framework based on dual-subspace orthogonal projection, where the suppress component is purified to attenuate the undesired semantics without compromising the essential features of the desired subject and scene. Our simple yet effective training-free method enables precise semantic control within the text embedding space, paving the way for more robust and faithful visual storytelling applications.
\section{Acknowledgements}

This work was supported by Institute of Information \& Communications Technology Planning \& Evaluation(IITP) grant funded by the Korea government(MSIT) (No. RS-2024-00439762, Developing Techniques for Analyzing and Assessing Vulnerabilities, and Tools for Confidentiality Evaluation in Generative AI Models).
{
    \small
    \bibliographystyle{ieeenat_fullname}
    \bibliography{main}
}


\end{document}